\documentclass[10pt,twocolumn,letterpaper]{article}

\usepackage[pagenumbers]{cvpr} 

\definecolor{cvprblue}{rgb}{0.21,0.49,0.74}
\usepackage[pagebackref,breaklinks,colorlinks,allcolors=cvprblue]{hyperref}

\usepackage[export]{adjustbox}
\usepackage{multirow,multicol}
\usepackage{colortbl}
\usepackage{makecell}
\usepackage{tcolorbox}
\definecolor{mygray}{gray}{0.6}

\def\paperID{} 
\def\confName{CVPR}
\def\confYear{2026}

\newcommand{\homepage}{\raisebox{-1.5pt}{\includegraphics[height=1em]{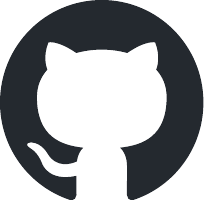}}}
\newcommand{\hfmodel}{\raisebox{-1.5pt}{\includegraphics[height=1em]{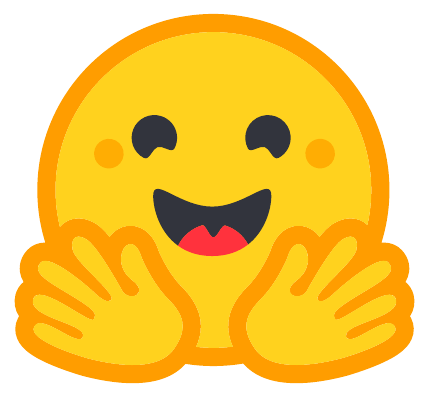}}}

\title{
ThinkGen: Generalized Thinking for Visual Generation
}

\author{Siyu Jiao$^{1*}$
\
Yiheng Lin$^{1*}$
\
Yujie Zhong$^{2\dag}$
\
Qi She$^{2}$
\
Wei Zhou$^{2}$
\
Xiaohan Lan$^{2}$  \\
\
Zilong Huang$^{2}$
\
Fei Yu$^{2}$
\
Yingchen Yu$^{2}$
\
Yunqing Zhao$^{2}$
\
Yao Zhao$^{1}$
\
Yunchao Wei$^{1\dag}$
\\\\[-3pt]
$^{1}$ Beijing Jiaotong University  ~~~~~~  $^{2}$ Bytedance 
\\\\[-3pt]
{\homepage ~ \texttt{Home: \!\!\!\!\!\url{https://github.com/jiaosiyuu/ThinkGen}}} \\
{\hfmodel ~ \texttt{HF: \!\!\url{https://huggingface.co/JSYuuu/ThinkGen}}} \\
}

\begin{document}
\maketitle

\makeatletter
\renewcommand*{\@makefnmark}{}
\footnotetext{
$^*$Equal Contribution. $^\dag$ Corresponding authors. 
}

\begin{abstract}

Recent progress in Multimodal Large Language Models (MLLMs) demonstrates that Chain-of-Thought (CoT) reasoning enables systematic solutions to complex understanding tasks. However, its extension to generation tasks remains nascent and limited by scenario-specific mechanisms that hinder generalization and adaptation. In this work, we present ThinkGen, the first think-driven visual generation framework that explicitly leverages MLLM's CoT reasoning in various generation scenarios. 
ThinkGen employs a decoupled architecture comprising a pretrained MLLM and a Diffusion Transformer (DiT), wherein the MLLM generates tailored instructions based on user intent, and DiT produces high-quality images guided by these instructions.
We further propose a separable GRPO-based training paradigm (SepGRPO), alternating reinforcement learning between the MLLM and DiT modules. This flexible design enables joint training across diverse datasets, facilitating effective CoT reasoning for a wide range of generative scenarios.
Extensive experiments demonstrate that ThinkGen achieves robust, state-of-the-art performance across multiple generation benchmarks.

\end{abstract}    
\section{Introduction}
\label{sec:intro}

Recent advances in Large Language Models (LLMs) \cite{deepseekmath, yang2025qwen3, xiaomi2025mimo} and Multimodal Large Language Models (MLLMs) \cite{openAI2025gpt4o, guo2025deepseek, qwen3vl} have demonstrated the effectiveness of Chain-of-Thought (CoT) reasoning, where models generate explicit intermediate steps to systematically solve complex tasks. CoT reasoning has significantly improved performance in areas such as mathematics, coding, and vision-language understanding. Building on these successes, researchers are now increasingly exploring how CoT reasoning can be leveraged to enhance generation tasks.

\begin{figure}[t]
\begin{center}
   \includegraphics[width=0.99\linewidth]{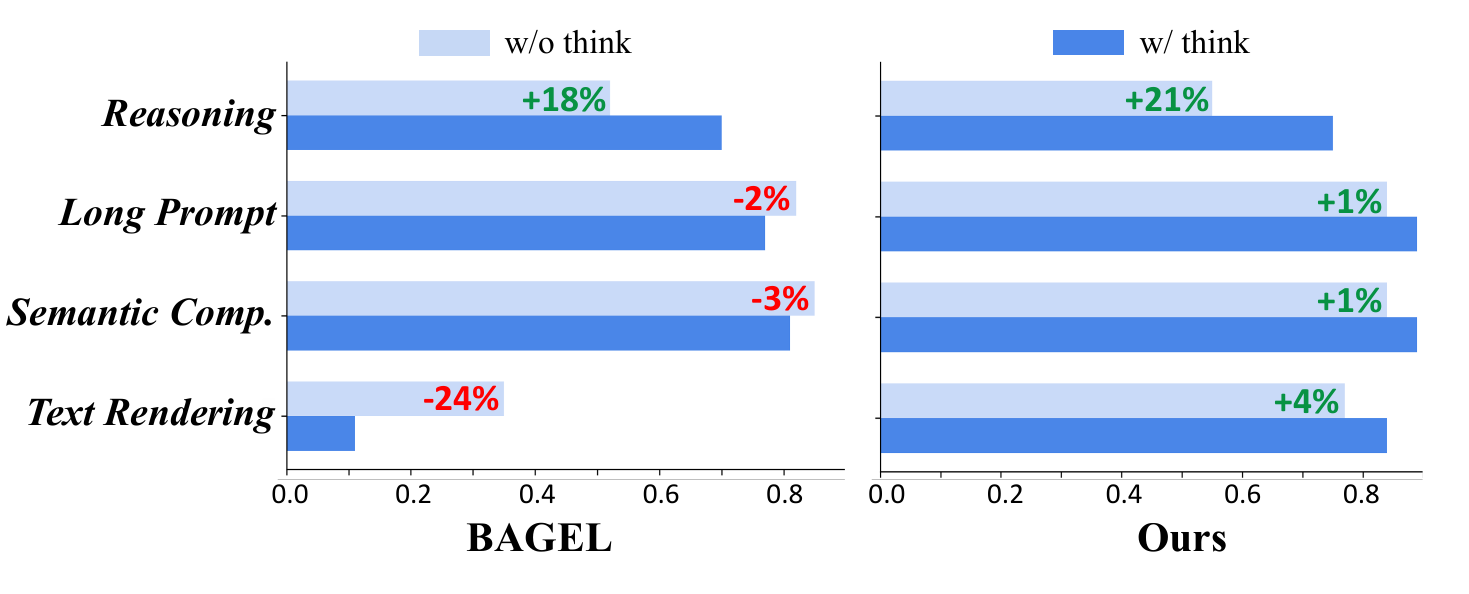}
\end{center}
    \vspace{-7mm}
   \caption{
   Comparison between BAGEL \cite{deng2025emerging} and our ThinkGen. ThinkGen achieves superior performance when adopting CoT reasoning (w/ think) across a wide range of generation scenarios.
    }
\label{fig:intro}
\end{figure}

\begin{figure*}[t]
\begin{center}
    \vspace{-4mm}
   \includegraphics[width=0.99\linewidth]{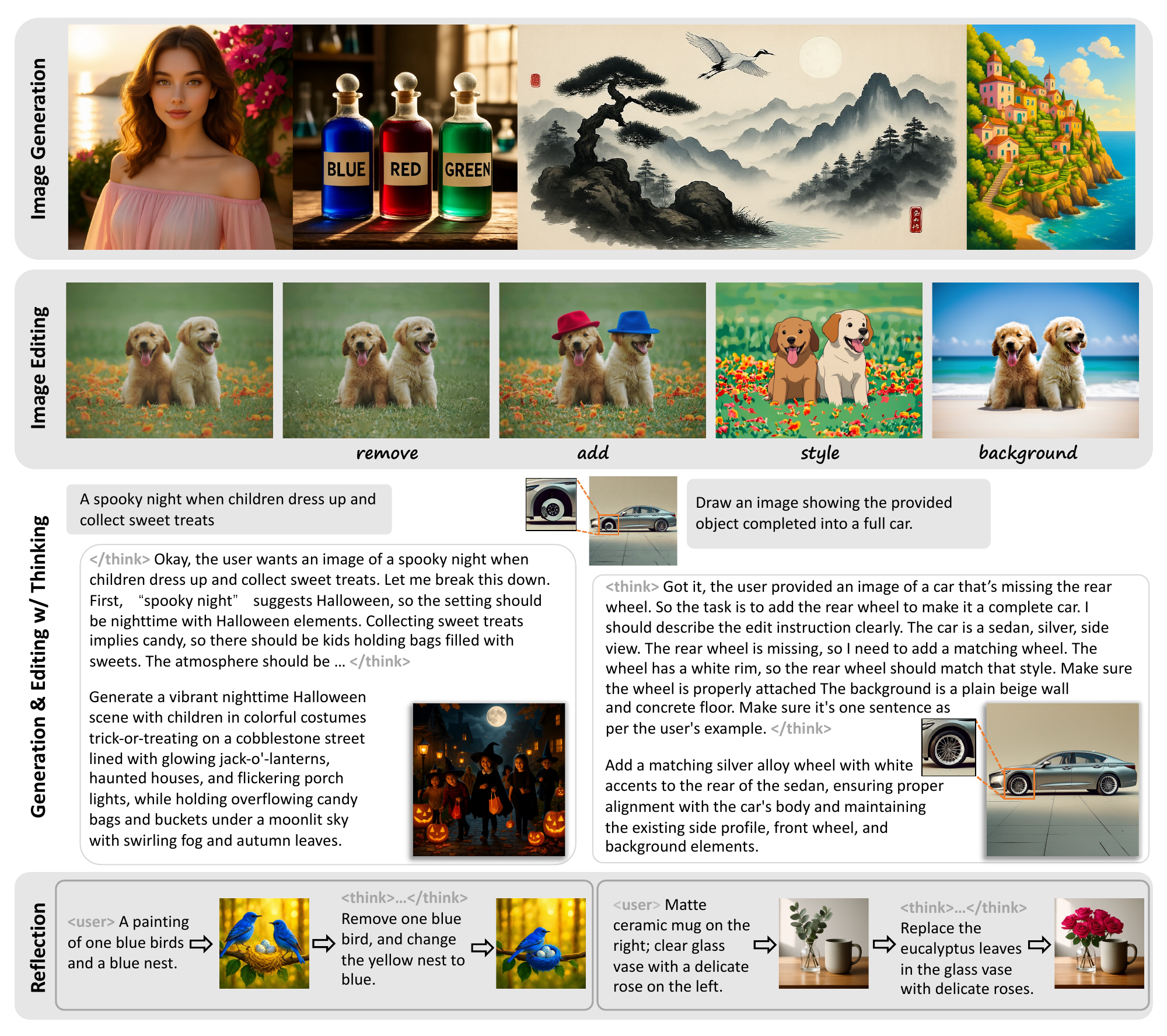}
\end{center}
    \vspace{-5mm}
   \caption{
    ThinkGen enables think-driven generation across a wide range of scenarios, including text-to-image generation, text rendering, image editing, reasoning generation, reasoning editing, and reflection.
   }
\label{fig:teaser}
\end{figure*}

Currently, CoT for generation remains at a preliminary stage. Pioneering work \cite{guo2025can} conceptualizes the progressive generation of image tokens as a form of CoT like textual token generation, and focuses on optimizing this process. Recent studies \cite{wang2025mint, deng2025emerging, jiang2025t2i, qin2025uni} advance the field by refining generation instructions \cite{deng2025emerging} or decomposing the generation process into distinct steps  \cite{wang2025mint, jiang2025t2i, qin2025uni}, thereby improving image quality in specific tasks. Despite these advances, current methods are constrained by a significant challenge: their CoT mechanisms are typically tailored to a single scenario, \textit{e.g.}, reasoning generation, and may degrade performance when applied to broader tasks (Fig. \ref{fig:intro} left). As a result, these approaches typically require manual intervention to activate CoT reasoning for different generation tasks, preventing their flexibility across diverse scenarios.

We attribute the aforementioned challenges to the fact that current frameworks often lack advanced reasoning capabilities.
In this work, we introduce ThinkGen, the first think-driven visual generation framework that explicitly leverages a Multimodal Large Language Model (MLLM) with $\texttt{<think>}$ formatting, endowing the system with robust reasoning abilities. This is followed by a dedicated Diffusion Transformer (DiT) for high-quality image synthesis. 
A key challenge lies in filtering out redundant information from the chain-of-thought (CoT) reasoning process to make it suitable for guiding the DiT. To this end, we introduce the Visual Generation Instruction refinement (VGI-refine) module, which extracts concise instruction information from the MLLM’s reasoning chain and concatenates it with learnable Prepadding States. This enables adaptive adjustment of the MLLM’s representation distribution, ensuring better alignment with the requirements of the DiT.

Our training paradigm combines supervised learning with reinforcement learning.
In the supervised learning stage, we develop a data-templete to generate pseudo-CoT annotations from image-text pairs, addressing the lack of explicit $\texttt{<think>}$ labels in most existing generation datasets, enabling the DiT to be optimized in a reasoning-driven manner.
In the RL stage, we introduce a separable GRPO-based training paradigm (SepGRPO), where GRPO is applied separately to the MLLM and DiT modules.
SepGRPO first freezes DiT while optimizing the MLLM, and then reverses the process by training DiT with the MLLM held fixed. 
To enhance generalization, we incorporate multi-scenario training data, jointly training the entire model across diverse datasets to achieve robust CoT reasoning in a wide range of generation tasks.
By performing the separable design, several advantages are provided:
1) Flexible Reward Design: Distinct rewards can be tailored for each module, enabling more targeted and effective optimization.
2) Reduced Learning Complexity: The MLLM focuses on providing instructions that are well-aligned with DiT’s preferences, while DiT specializes in producing high-quality images based on these tailored instructions.
3) Lower Training Cost: The separate design significantly reduces GPU memory usage during training, greatly enhancing computational efficiency.

We evaluate ThinkGen across various generation scenarios. Extensive experiments demonstrate that ThinkGen achieves robust performance on diverse generation benchmarks, \textit{e.g.}, GenEval (0.89), CVTG (0.84), and ImgEdit (4.21).
Notably,  enabling CoT reasoning in ThinkGen yields substantial improvements on reasoning benchmarks: WISE: 0.55$\rightarrow$0.76, RISEBench: 3.6$\rightarrow$13.0.


\begin{figure*}[t]
\begin{center}
\vspace{-2mm}
   \includegraphics[width=0.99\linewidth]{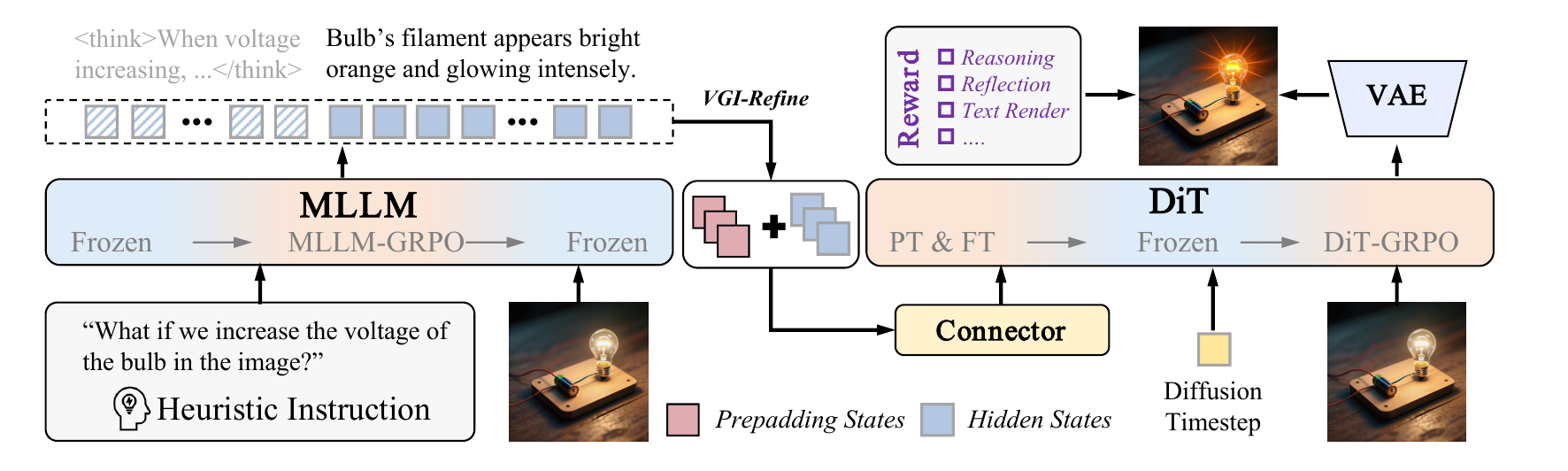}
\end{center}
    \vspace{-4mm}
   \caption{
    \textbf{Overview of ThinkGen.} Within ThinkGen, the MLLM and DiT architectures are decoupled for autoregressive CoT generation and diffusion-based image generation. The MLLM receives text/images as input and outputs generation instructions tailored to the preferences of the DiT. Through a process called visual generation instruction refinement (VGI-Refine), the hidden states corresponding to these instructions are extracted and concatenated with Prepadding States, forming the conditional information for DiT's image generation. For clarity, we omit the text encoder and vision encoder components within MLLM and DiT.
   }
\label{fig:framework}
\end{figure*}

\section{Related Work}
\label{sec:related}

\noindent \textbf{Unified Model for Generation and Understanding.}
Recently, building unified models for both generation and understanding has attracted significant attention. Leveraging the strong multimodal understanding capabilities of MLLMs across images and text, image generation performance has seen further improvements.

One line of work \cite{wu2025janus, chen2025janus, wang2024emu3} adopts VQGAN-style tokenizers \cite{esser2021taming} and trains MLLMs to generate discrete visual tokens, producing images via next-token prediction in an autoregressive manner. \cite{wu2025omnigen2, lin2025uniworld, chen2025blip3, pan2025transfer, wu2025qwen} integrates MLLMs with text-to-image diffusion models \cite{podell2023sdxl, labs2023flux}. The powerful MLLMs is used to extract semantic features, which are then fed as conditions to a diffusion model for image generation. \cite{wu2025omnigen2} uses the last hidden states as conditional features for generation, while \cite{pan2025transfer} introduces learnable queries to extract informative features for conditioning. However, these methods primarily treat the MLLM as a feature extractor, without fully leveraging its CoT reasoning capabilities. Additionally, some works \cite{deng2025emerging, ma2025janusflow, xie2025show, cao2025hunyuanimage30technicalreport} fuse autoregressive and diffusion modeling. This paradigm autoregressively generates text tokens while producing image tokens via a multi-step diffusion process, combining the strengths of both approaches.

\noindent \textbf{Reinforcement Learning.} 
Recently, reinforcement learning (RL) has been used to enhance MLLMs and diffusion-based generative models. Online RL \cite{schulman2017proximal, guo2025deepseek, jaech2024openai} for MLLMs has been effective at improving MLLMs reasoning capabilities and aligning outputs with human preferences. In particular, \cite{guo2025deepseek} shows that rule-based reward functions can elicit human-like, complex chain-of-thought reasoning, while also being memory-efficient by removing the need for a separate value model. A number of works \cite{liu2025flow, luo2025editscore, xue2025dancegrpo} also apply GRPO to flow-matching models \cite{esser2024scaling, labs2023flux} with task-specific rewards. This yields a stable approach for aligning visual outputs with human preferences, improving aesthetics, text rendering, and image–prompt consistency.

\section{Model Architecture}
\label{sec:architecture}

We introduce ThinkGen, a think-driven unified model designed for various visual generation tasks, with its architecture shown in Fig.~\ref{fig:framework}. Our model utilizes decoupled MLLM and DiT modules, dedicated to understanding and generation, respectively. This design ensures optimal performance for each component while maintaining both scalability and modularity within the system. For generation tasks, the MLLM receives an image caption or reference image(s) along with editing instructions as input, and outputs rewritten generation instructions tailored to the preferences of DiT. The DiT module then uses these refined instructions to generate high-quality images.

\subsection{Multimodal Large Language Model}

As shown in Fig. \ref{fig:framework}, ThinkGen leverages a MLLM to process both visual and textual inputs, employing autoregressive generation for CoT reasoning. The MLLM is initialized with Qwen3-VL-8B-Think \cite{qwen3vl}.
For image generation tasks, we design a specialized system prompt ($\texttt{[SYS]}$) to encourage the MLLM to understand user intent and provide appropriate rewrite instructions. We then extract the final two layers of hidden states generated after the $\texttt{</think>}$ token as conditional inputs for DiT. Empirical results indicate that using the last two layers of hidden states significantly benefits visual generation.

\subsection{Diffusion Transformer}

ThinkGen employs a standard DiT architecture \cite{wu2025omnigen2, lin2025uniworld} initialized with OmniGen2-DiT-4B \cite{wu2025omnigen2}, where the output from the MLLM is used as conditional textual input for generation. In image edit task, additional reference image(s) are processed by a VAE \cite{van2017neural} and incorporated as conditional visual inputs. The visual and textual inputs are concatenated with the noisy latent features, enabling joint attention across modalities.
We employ a simple linear layer as a connector to align features from multiple conditional inputs. We experimentally find that this straightforward linear projection outperforms MLP-based or more complex transformer-based connectors.

\noindent \textbf{VGI-refine.}
To address the redundancy in the MLLM's autoregressive chain-of-thought (CoT) outputs \cite{Xu2025Chain, Xia2502TokenSkip}, we introduce Visual Generation Instruction Refinement (VGI-refine), which consists of two steps. First, instruction tokens following the special token $\texttt{</think>}$ are extracted from the text tokens generated by the MLLM, thereby isolating the essential CoT results for downstream image generation. Second, we concatenate $K$ learnable Prepadding States to the extracted instruction tokens. This concatenation regulates the data distribution of the output hidden states and is especially beneficial for short instructions (\textit{e.g.}, \textit{generate a dog} or \textit{remove the cat}). The resulting refined instruction states are then provided as conditional input to the DiT.

\section{Training Recipe}
\label{sec:training}

\begin{figure*}[t]
\begin{center}
    \vspace{-3mm}
   \includegraphics[width=0.99\linewidth]{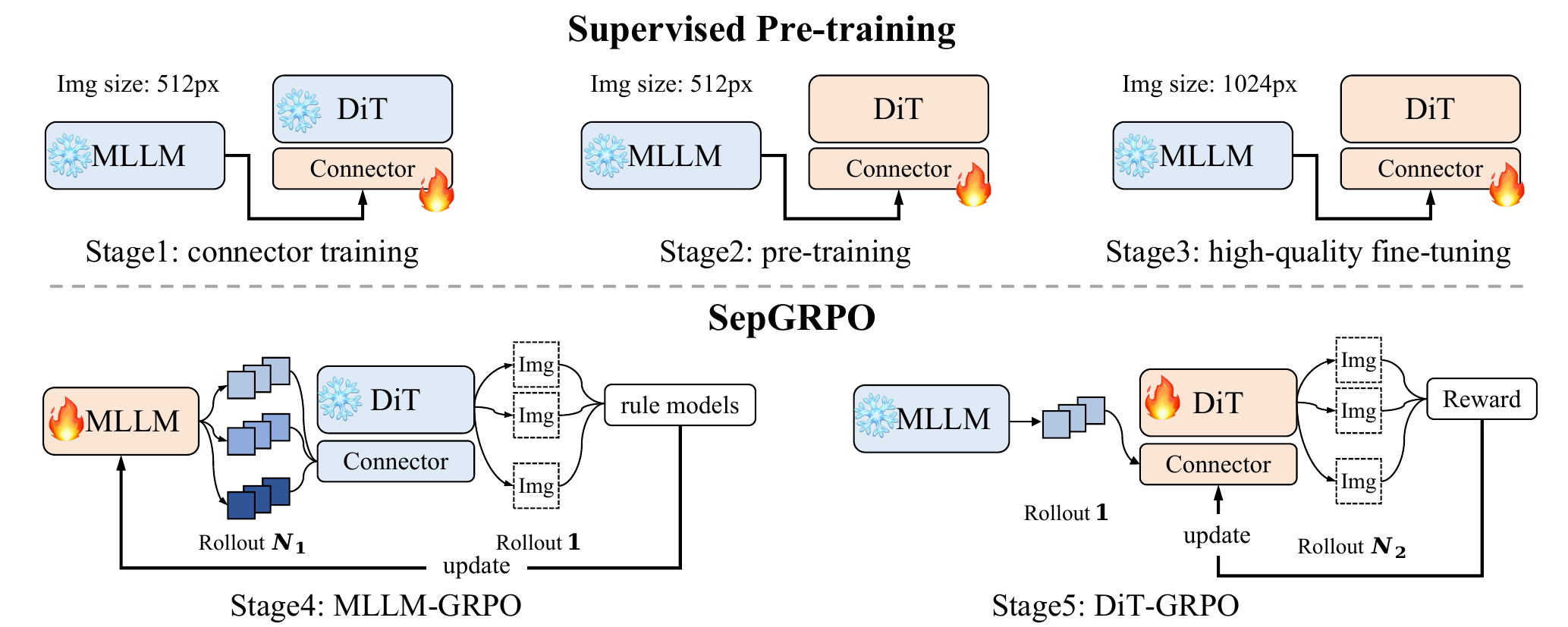}
\end{center}
   \caption{
    The training recipe of ThinkGen consists of three supervised pre-training stages: Connector training (stage 1), Pre-training (stage 2), and High-quality fine-tuning (stage 3), as well as two SepGRPO stages: MLLM-GRPO (stage 4) and DiT-GRPO (stage 5).
   }
\label{fig:training}
\end{figure*}

Our ThinkGen training is divided into five distinct stages. Initially, we perform Supervised Pre-training on DiT (Stage 1–3) to ensure high-quality image generation. Subsequently, we introduce a separable MLLM and DiT reinforcement learning approach called SepGRPO (Stage 4–5). Through SepGRPO training, the MLLM learns to generate captions or editing instructions that are optimally aligned with DiT’s preferences, while DiT is further refined to produce superior images based on these tailored instructions. The overall training workflow is depicted in Fig. \ref{fig:training}.

\subsection{Supervised Pre-training}
\label{sec:pretraining}

The Supervised Pre-training stages (Stage 1–3) are designed to align the DiT with the MLLM, while simultaneously enhancing image generation quality.  We adopt the Rectified Flow \cite{liu2022flow} training paradigm, which directly regress the velocity field $v_\theta(x_t, t)$ by minimizing the Flow Matching objective~\cite{lipman2022flow,liu2022flow}:
\begin{equation}
    \mathcal{L}(\theta) = \mathbb{E}_{t,\, x_0 \sim X_0,\, x_1 \sim X_1} \left[ \left\| \mathbf{v} - v_\theta(x_t, t) \right\|^2 \right],
    \label{eq:fm-loss}
\end{equation}
here $\mathbf{v} = x_1 - x_0$ denotes the target velocity field.

\noindent \textbf{Input Format.} 
Rewriting each caption or edit instruction during pre-training would be prohibitively expensive. Therefore, in Stage 1–3, we construct pseudo-CoT templates to simulate the MLLM’s CoT process. Specifically, we leave the content within $\texttt{<think> </think>}$ empty and simply repeat the original caption or edit instruction as the answer. The resulting template is: $\texttt{[SYS]+[C]+<think> </think>+[C]}$, where $\texttt{[SYS]}$ denotes the system prompt, and $\texttt{[C]}$ denotes the image caption or editing instruction.

\noindent \textbf{Stage1 Alignment.} 
In this stage, we introduce $K$ Learnable prepadding states and align the DiT with the MLLM by training only the linear connector, while keeping the MLLM and DiT frozen. Each image is resized to $\leq$ 512$\times$512px.

\noindent \textbf{Stage2 Pre-training.} 
During this stage, all DiT parameters are trainable. The training corpus comprises 60M image samples, consisting of text-to-image, image edit, text rendering and in-context generation data.  Each image is resized to no more than 512$\times$512 pixels.

\noindent \textbf{Stage3 High-quality fine-tuning.} 
In the supervised fine-tuning stage, we construct a 0.7M high-quality subset to enhance DiT’s instruction-following capability and image aesthetic. The maximum of training resolution is set to 1024$\times$1024 pixels.

\subsection{SepGRPO}
We propose SepGRPO, an RL training strategy designed to encourage the MLLM to generate captions/editing instructions that are optimally aligned with DiT’s preferences, while enabling DiT to produce higher-quality images based on these instructions. SepGRPO decouples the rollout process for text and vision: first, DiT is fixed while GRPO is applied to the MLLM through joint multi-task training; then, the MLLM is fixed while GRPO is applied to DiT.

\noindent \textbf{Input Format.} 
We design a specialized $\texttt{[SYS]}$ during on-policy training to facilitate a cold start, allowing the MLLM to explore text conditions favored by DiT. Specifically, We concatenate the $\texttt{[SYS]}$, the input sample $\texttt{[C]}$, and a special $\texttt{<think>}$ token as the input to the MLLM. The resulting template is: $\texttt{[SYS]+[C]+<think>}$.

\noindent \textbf{Stage4 MLLM-GRPO.} 
In this stage, we apply GRPO to the MLLM to encourage the generation of rewritten text that aligns with DiT’s preferences. We optimize the MLLM on multiple scenarios to enhance the generalization capability of CoT reasoning. Specifically, we select five representative generation scenarios: semantic composition, reasoning generation, text rendering, image editing, and reflection.
For each scenario, we collect and curate dedicated datasets and design corresponding rule models to guide the optimization. The details of the datasets and rule models for each scenario are summarized in Table \ref{tab:scenario}.

\begin{table}[h]
    \centering
    \resizebox{0.99\linewidth}{!}{
    \begin{tabular}{l|l|l}
    \toprule
    \textbf{Scenario} & \textbf{Dataset} & \textbf{Rule Model} \\
    \midrule
    Semantic composition & 5K semantic prompts & GenEval \cite{ghosh2023geneval} \\
    Reasoning generation & 10K reasoning prompts & HPSv3 \cite{ma2025hpsv3} \\
    Text rendering & 3K text rendering prompts & Word Acc. \cite{cvtg} \\
    Image editing & 3K editing samples & SigLIP2 \cite{tschannen2025siglip} \\
    Reflection & 3K reflection samples &  NED \\
    \bottomrule
    \end{tabular}
    }
    \caption{Training data and Rule Models in MLLM-GRPO. Notably, all training data and evaluation benchmarks are strictly non-overlapping, ensuring unbiased assessment.}
    \label{tab:scenario}
\end{table}

For each input to the MLLM, we perform $N_1$ rollouts from the policy $\pi_{\theta_\text{old}}$ to generate trajectories $\{ o_i\}_{i=1}^{N_1}$, which are subsequently used by DiT to produce the corresponding images. Specifically, DiT generates one image for each trajectory. To mitigate the impact of image generation stochasticity, we ensure that all trajectories corresponding to the same input share identical latent noise. The corresponding rule models are then used to calculate a reward $\mathcal{R}_i$ for each trajectory. Subsequently the advantage $\hat{A}_{i}$ for the $i$-th trajectory is computed in a group-relative manner:
\begin{equation}
\label{eq:advantage}
\hat{A}_{i} = \frac{\mathcal{R}_i - \text{mean}(\{\mathcal{R}_i\}_{i=1}^{N_1})}{\text{std}(\{\mathcal{R}_i\}_{i=1}^{N_1})}.
\end{equation}
The policy $\pi_{\theta_\text{old}}$ is then updated by optimizing the GRPO objective, which is a clipped surrogate function with KL-divergence regularization:
{\scriptsize
\begin{equation}
\begin{aligned}
\mathcal{J}_\text{GRPO}(\theta)& = \mathbb{E}_{(q,a)\sim \mathcal{D}, \{o_i\}_{i=1}^G\sim \pi_{\theta_\text{old}}(\cdot\mid q)} \\
&\Bigg[ \frac{1}{\sum_{i=1}^{G}|o_i|}\sum_{i=1}^{G}\sum_{t=1}^{|o_i|} \Bigg( \mathrm{MIN}
 - \beta D_{\text{KL}}(\pi_{\theta} || \pi_{\text{ref}}) 
\Bigg) \Bigg]
\label{eq:grpoloss}
\end{aligned}
\end{equation}
}
{\scriptsize
\begin{equation}
\mathrm{MIN}  = \text{min} \Big( r_{i,t}(\theta) \hat{A}_{i},
\mathrm{CLIP}\Big( r_{i,t}(\theta), 1 - \varepsilon, 1 + \varepsilon \Big)\Big),
\end{equation}
}
where $r_{i,j}(\theta)$ denotes the ratio between the probabilities of $\pi_\theta$ and $\pi_{\theta_{\text{old}}}$ for outputting the current token. 

In this process, DiT and the rule models jointly serve as reward models. This diverse reward design allows our model to adaptively apply CoT reasoning across a wide range of generation tasks.
We provide detailed descriptions of the $\texttt{[SYS]}$, training data distribution, and rule model settings in the appendix.

\noindent \textbf{Stage5 DiT-GRPO:} In this stage, we apply FlowGRPO \cite{liu2025flow} to enhance the instruction-following capability of DiT. We utilize data from the \textit{Simple Scene} and \textit{Text Rendering} scenarios, along with their corresponding reward calculation methods. The training data used in this stage is strictly non-overlapping with that of Stage 4. For each input, the frozen MLLM first performs a single rollout to generate a CoT reasoning trajectory, after which DiT conducts $N_2$ rollouts to generate $N_2$ corresponding images. We then compute the advantages as defined in Equation \ref{eq:advantage} and update the DiT's policy by maximizing the GRPO objective in Equation \ref{eq:grpoloss}. This process encourages the DiT to favor trajectories that yield higher rewards.

\noindent \textbf{Denoising Reduction:} 
Denoising Reduction \cite{liu2025flow} (20 steps with 512px) is employed to accelerate the sampling process. This approach enables the efficient collection of low-quality yet informative trajectories during training.

\begin{table*}[t]
    \centering
    \scriptsize
    \vspace{-4mm}
    \resizebox{0.8\linewidth}{!}{
    \begin{tabular}{l|cccccc|c}
    \toprule
    \textbf{Model}  & \textbf{Cultural} & \textbf{Time}  & \textbf{Space} & \textbf{Bioligy} & \textbf{Physics} & \textbf{Chemistry}  & \textbf{Overall} \\
    \midrule
    GPT-4o \cite{chen2025sharegpt4oimg} & 0.81	& 0.71	& 0.89	& 0.83 & 0.79& 0.74& 0.80 \\
    \midrule
    \multicolumn{8}{c}{\textbf{\textit{Gen. Only}}} \\   
    SDXL~\citep{podell2023sdxl}  & 0.43 & 0.48 & 0.47 & 0.44 & 0.45 & 0.27 & 0.43 \\
    SD-3.5-large \cite{esser2021taming} & 0.44 & 0.50 & 0.58 & 0.44 & 0.52 & 0.31 & 0.46 \\
    FLUX.1-dev~\citep{labs2023flux}  & 0.48 & 0.58 & 0.62 & 0.42 & 0.51 & 0.35 & 0.50 \\
    PixArt-$\alpha$~\citep{chen2024pixart}  & 0.45 & 0.50 & 0.48 & 0.49 & 0.56 & 0.34 & 0.47\\
    \midrule
    \multicolumn{8}{c}{\textbf{\textit{Und. and Gen.}}} \\
VILA-U~\citep{wu2024vila}  & 0.26 & 0.33 & 0.37 & 0.35 & 0.39 & 0.23 & 0.31\\
Janus-Pro-7B~\citep{chen2025janus} & 0.30 & 0.37 & 0.49 & 0.36 & 0.42 & 0.26 & 0.35 \\
Emu3 \cite{wang2024emu3} & 0.34 & 0.45 & 0.48 & 0.41 & 0.45 & 0.27 & 0.39 \\
Show-o~\citep{xie2025show} & 0.28 & 0.40 & 0.48 & 0.30 & 0.46 & 0.30 & 0.35 \\
MetaQuery-XL~\citep{pan2025transfer} & 0.56 & 0.55 & 0.62 & 0.49 & 0.63 & 0.41 & 0.55 \\
BLIP3-o-8B~\citep{chen2025blip3} & -- & -- & -- & -- & -- & -- & 0.62 \\
BAGEL~\citep{deng2025emerging}& 0.44 & 0.55 & 0.68 & 0.44 & 0.60 & 0.39 & 0.52\\
BAGEL*~\citep{deng2025emerging} & 0.76 & 0.69 & 0.75 & 0.65 & 0.75 & 0.58 & 0.70 \\
OmniGen2~\citep{wu2025omnigen2} & 0.42& 0.52 & 0.64 & 0.43 & 0.50 & 0.34 & 0.47\\
STAR~\citep{star} & 0.61& 0.67 & 0.61 & 0.74 & 0.69 & 0.66 & 0.66\\

    \midrule 
    \textbf{ThinkGen}  & 0.53 & 0.55 & 0.71 & 0.51 & 0.58 & 0.40 & 0.55 \\
    \textbf{ThinkGen*} & \textbf{0.78} &  \textbf{0.73} & \textbf{0.85} & \textbf{0.74} & \textbf{0.74} & \textbf{0.68} & \textbf{0.76}\\
     
    \bottomrule
    \end{tabular}
    }
    \caption{Evaluation of reasoning generation ability on WISE benchmark.  * denotes that CoT reasoning is utilized during image generation.}
    \label{tab:wise}
\end{table*}

\section{Experiments}
\label{sec:exp}
In this section, we first provide a brief overview of the data composition (Sec. \ref{sec:data}) and evaluation setup (Sec. \ref{sec:evaluation}). Next, we evaluate ThinkGen across a variety of visual generation benchmarks (Sec. \ref{sec:comparison}). Furthermore, we conduct detailed ablation studies to verify the contribution of each component and training strategy (Sec. \ref{sec:abs}). We also analyze the SepGRPO process (Sec. \ref{sec:grpo}) .

\subsection{Data composition}
\label{sec:data}
For text-to-image generation, our training dataset comprises 54M image-text pairs sourced from publicly available datasets \cite{chen2025sharegpt4oimg, david_beniaguev_2024_SFHQ_T2I, fang2025flux, wang2019densefusion, chen2025blip3}.  For image editing tasks, we utilize a diverse set of open-source image editing datasets \cite{zhao2024ultraedit, song2025omniconsistency, ye2025echo4o, wang2025gpt, chen2025sharegpt4oimg, ma2025x2editrevisitingarbitraryinstructionimage, Layer2025NoHumansRequired}, totaly 5M samples. Furthermore, 1M high-quality proprietary samples are used to further enhance the model's ability to generate visually appealing images and text rendering ability, see the appendix for more details.

\subsection{Evaluation setup}
\label{sec:evaluation}

\noindent \textbf{Reasoning Generation.} 
We assess reasoning generation capability on WISEBench \cite{niu2025wise}, a world knowledge-informed semantic evaluation benchmark (1000 prompts). 

\noindent \textbf{Reasoning Editing.} 
Reasoning editing capability is evaluated on RISEBench  \cite{risebench} (360 pairs). RISEBench evaluates the model’s reasoning editing capability across four fundamental types: temporal reasoning, causal reasoning, spatial reasoning, and logical reasoning.

\noindent \textbf{Text-to-image Evaluation.} 
This task evaluates semantic consistency on GenEval \cite{ghosh2023geneval} (553 prompts) and long-form generation ability on DPG-Bench \cite{dpg} (1065 prompts), as well as text rendering capability on CVTG \cite{cvtg} (2000 prompts). In the CVTG benchmark, we report the word accuracy of text rendering to assess model performance.

\noindent \textbf{Image Editing Evaluation.} 
We assess image editing capability on ImgEdit \cite{ye2025imgedit} (737 pairs), which covers object-level, background, style, and composite manipulations.

\begin{table}[h]
    \scriptsize
    \centering
    \resizebox{0.99\linewidth}{!}{
    \begin{tabular}{l|cccc|c}
    \toprule
    \textbf{Models} & \textbf{Tem.} & \textbf{Cau.} & \textbf{Spa.} & \textbf{Log.} & \textbf{Avg.} \\
    \midrule
    \multicolumn{6}{c}{\textbf{\textit{Closed-source}}} \\   
GPT-4o~\cite{chen2025sharegpt4oimg} & 34.1 & 32.2 & 37.0 & 10.6 & 28.9 \\
Gemini-2.0~\cite{team2023gemini} & 8.2 & 15.5 & 23.0 & 4.7 & 13.3 \\
    \midrule
    \multicolumn{6}{c}{\textbf{\textit{Open-source}}} \\

OmniGen~\cite{xiao2025omnigen} & 1.2 & 1.0 & 0.0 & 1.2 & 0.8 \\
EMU2~\cite{emu2} & 1.2 & 1.1 & 0.0 & 0.0 & 0.5 \\
Step1X-Edit~\cite{liu2025step1x} & 0.0 & 2.2 & 2 & 3.5 & 1.9 \\
HiDream-Edit~\cite{HiDream.ai.}~~~~~ & 0.0 & 0.0 & 0.0 & 0.0 & 0.0 \\
FLUX-Canny~\cite{flux1-canny} & 0.0 & 0.0 & 0.0 & 0.0 & 0.0 \\
BAGEL~\cite{deng2025emerging} & 3.5 & 4.4 & 9.0 & 5.9 & 5.8 \\
BAGEL*~\cite{deng2025emerging} & 5.9 & 17.8 & 21.0 & 1.2&11.9 \\
OmniGen2~\citep{wu2025omnigen2} & 0.0& 2.2 & 7.0 & 2.3  & 3.0\\

    \midrule 
    \textbf{ThinkGen} & 3.5 & 2.2 &  7.0& 1.1 & 3.6 \\
    \textbf{ThinkGen*} & \textbf{16.4} &  \textbf{17.7} & \textbf{16.0} & \textbf{1.1} & \textbf{13.0}\\
     
    \bottomrule
    \end{tabular}
    }
    \caption{Evaluation of reasoning editing ability on RISEBench.}
    \label{tab:risebench}
\end{table}

\subsection{Comparison with the state of the art methods}
\label{sec:comparison}

\begin{table*}[t]
    \scriptsize
    \centering
    \vspace{-3mm}
    \resizebox{0.99\linewidth}{!}{
    \begin{tabular}{l|ccc|ccccc|cc}
    \toprule
     & \multicolumn{3}{c|}{\textbf{GenEval}} & \multicolumn{5}{c|}{\textbf{DPG}}  & \multicolumn{2}{c}{\textbf{CVTG}} \\
    \textbf{\multirow{-2}{*}{Model}} & \textbf{Counting} &\textbf{Position} & \textbf{Overall} & \textbf{Global} & \textbf{~Entity} & \textbf{Attribute} & \textbf{Relation} & \textbf{Overall} & \textbf{Acc.}& \textbf{NED}\\
    \midrule
    \multicolumn{11}{c}{\textbf{\textit{Gen. Only}}} \\   
    SDXL~\citep{podell2023sdxl}  & 0.39& 0.15& 0.55& 83.27 & 82.43 & 80.91 & 86.76 & 74.65 & - & -\\
    FLUX.1-dev~\citep{labs2023flux}  & 0.75& 0.68& 0.82& 82.10 & 89.50 & 88.70 & 91.10 & 84.00 & 0.49 & 0.68 \\
    PixArt-$\alpha$~\citep{chen2024pixart}  & 0.44 & 0.08 & 0.48 & --& -- & --& -- & --& --\\
    SD3-Medium~\citep{esser2024scaling}  & 0.72& 0.33& 0.74& 87.90 & 91.01 & 88.83 & 80.70 & 84.08 & 0.65 & 0.84\\
    Sana-1.6B~\citep{xie2024sana}  & 0.62 & 0.21 & 0.66 & --& -- &  --& -- & --& -- & -- \\
    TextCrafter~\citep{cvtg}  & -- & -- & -- & --& -- &  --& -- & --& 0.76 & 0.90 \\
    \midrule
    \multicolumn{11}{c}{\textbf{\textit{Und. and Gen.}}} \\
    Emu3-Gen~\citep{wang2024emu3} &0.34 &0.17& 0.54& 85.21 & 86.68 & 86.84 & 90.22 & 80.60 &--&--\\
    ILLUME+ ~\citep{huang2025illume+}& 0.62& 0.42& 0.72 & -- & -- & -- & -- & --\\
    Janus-Pro~\citep{chen2025janus}  &0.59 &0.79&0.80& 86.90 & 88.90 & 89.40 & 89.32 & 84.19  & -- & --\\
    MetaQuery-XL~\citep{pan2025transfer} ~~~~& --  & -- & 0.80 & -- & -- & -- & --  & 82.05 & -- & --\\
    BLIP3-o-8B~\citep{chen2025blip3} & -- & --  & 0.84 & -- & -- & -- & --  & 81.60 & -- & -- \\
    BAGEL~\citep{deng2025emerging} & 0.81& 0.64 &0.82 & 88.94 & 90.37 & 91.29 & 90.82 & 85.07 & 0.35 & 0.65 \\
    BAGEL*~\citep{deng2025emerging} & 0.78& 0.52 &0.79 & 90.13 & 90.41 & 88.73 & 88.22 & 83.46 & 0.11 & 0.39 \\
    OmniGen2~\citep{wu2025omnigen2} & 0.88 & 0.55 & 0.80 & 88.81 & 88.83 & 90.18 & 89.37 & 83.57 & 0.52 & 0.77\\
    \midrule 
    \textbf{ThinkGen}  & 0.81 & 0.79 & 0.88 & 90.32 & 90.86 & 91.23 & \textbf{92.48} & 85.14 & 0.80 & 0.91 \\
    \textbf{ThinkGen*} & \textbf{0.84} &  \textbf{0.80} & \textbf{0.89} & \textbf{90.87} & \textbf{91.36} & \textbf{91.77} & 91.52 & \textbf{85.87} & \textbf{0.84} & \textbf{0.94}\\
     
    \bottomrule
    \end{tabular}
    }
    \caption{Evaluation of text-to-image generation ability on GenEval, DPG and CVTG benchmark.}
    \label{tab:geneval}
\end{table*}
\begin{table}
    \scriptsize
    \centering
    \setlength{\tabcolsep}{2.5pt}
    \resizebox{0.99\linewidth}{!}{
    \begin{tabular}{l|cccccc|c}
    \toprule
    \textbf{Model}  & \textbf{Add} & \textbf{Adj.}  & \textbf{Rep.} & \textbf{Rem.} & \textbf{BG} & \textbf{Sty.}  & \textbf{Overall} \\
    \midrule
    GPT-4o~\citep{openAI2025gpt4o} & 4.61& 4.33& 4.35& 3.66& 4.57 &4.93 &4.20 \\
    \midrule
    \multicolumn{8}{c}{\textbf{\textit{Gen. Only}}} \\   
    MagicBrush~\citep{zhang2023magicbrush}  &2.84 &1.58& 1.97& 1.58& 1.75& 2.38 & 1.90\\
    Instruct-P2P~\citep{brooks2023instructpix2pix}  &  2.45& 1.83& 2.01& 1.50 &1.44& 3.55&  1.88 \\
    AnyEdit~\citep{yu2025anyedit}  &  3.18& 2.95  &2.47& 2.23 &2.24 &2.85& 2.45 \\
    UltraEdit~\citep{zhao2024ultraedit}  &  3.44 & 2.81& 2.96& 1.45& 2.83 &3.76 & 2.70\\
    Step1X-Edit~\citep{liu2025step1x}  & 3.88 & 3.14& 3.40& 2.41 &3.16& 4.63 & 3.06 \\
    ICEdit~\citep{zhang2025context}  & 3.58 &3.39& 3.15 &2.93& 3.08& 3.84&  3.05 \\
    \midrule
    \multicolumn{8}{c}{\textbf{\textit{Und. and Gen.}}} \\
    OmniGen~\citep{xiao2025omnigen}  &  3.47& 3.04 & 2.94& 2.43& 3.21& 4.19&   2.96 \\
    Janus-4o~\citep{chen2025sharegpt} & 3.60 & 3.25 & 3.27 & 2.28 & 3.32 & 4.47 &   3.26 \\
    BAGEL~\citep{deng2025emerging}  &  3.56 &3.31 &3.30& 2.62 &3.24& 4.49&   3.20 \\
     OmniGen2~\citep{wu2025omnigen2} & 3.57 & 3.06 & 3.74 & 3.20& 3.57& 4.81 &3.44 \\
    UniWorld~\citep{lin2025uniworld}  &  3.82& 3.64& 3.47& 3.24& 2.99& 4.21   &3.26 \\
   
    \midrule
      \textbf{ThinkGen} & 4.64 & 4.12 & 4.07 &  3.95  &  4.31 & 4.73 & 4.14\\
    \textbf{ThinkGen*}  & \textbf{4.75} & \textbf{4.25} &\textbf{4.15} & \textbf{3.49} & \textbf{4.3} & \textbf{4.68} & \textbf{4.21} \\

    \bottomrule
    \end{tabular}
    }
    \caption{Evaluation of image editing on ImgEdit benchmark.}
    \label{tab:imgedit}
\end{table}

We report results of ThinkGen \textit{w.} and \textit{w.o.} CoT reasoning. When generation \textit{w.o.} CoT reasoning, we simulate the CoT process by adopting the Input Format described in Sec. \ref{sec:pretraining}.

\noindent \textbf{Reasoning Generation.}
We conduct experiments on WISEBench to evaluate the reasoning generation capability. Tn Tab. \ref{tab:wise}, we compare ThinkGen with previous well-known generative models and unified generation-understanding models. Our ThinkGen demonstrates a significant advantage over methods based on direct generation. By leveraging CoT reasoning, ThinkGen achieves a substantial improvement of +21\% (0.55 $\rightarrow$ 0.76), and establishes a new state-of-the-art performance on WISEBench.

\noindent \textbf{Reasoning Editing.} 
As shown in Tab. \ref{tab:risebench}, on RISEBench, ThinkGen's CoT reasoning significantly surpasses open-source models (3.6$\rightarrow$13.0) and achieves results competitive with the closed-source model Gemini-2.0.

\noindent \textbf{Text-to-image Generation.} 
In Tab. \ref{tab:geneval}, we present the performance of ThinkGen on the GenEval, DPG-Bench, and CVTG benchmarks. With CoT reasoning, ThinkGen consistently demonstrates improvements across all scenarios, and achieves the best results among many well-known models. These results indicate that ThinkGen possesses strong instruction-following and text-rendering capabilities. 

\noindent \textbf{Image Editing.} 
In Tab. \ref{tab:imgedit}, we compare the performance of ThinkGen on ImgEdit. Compared with a range of open-source models, ThinkGen shows significantly superior metrics, achieving performance comparable to GPT-4o.

\subsection{Ablation Study}
\label{sec:abs}

\noindent \textbf{Training stage ablations.}
To understand the effect of each training stage in the ThinkGen, including the Supervised Pre-training and the SepGRPO. 
We start from a pretrained MLLM and DiT, and gradually apply each training stage (see Tab. \ref{tab:ab-stage}). We present results on GenEval, WISE, and CVTG, which are used to evaluate instruction-following, reasoning generation, and text-rendering, respectively.

\begin{figure*}[t]
    \vspace{-4mm}
\begin{center}
   \includegraphics[width=0.95\linewidth]{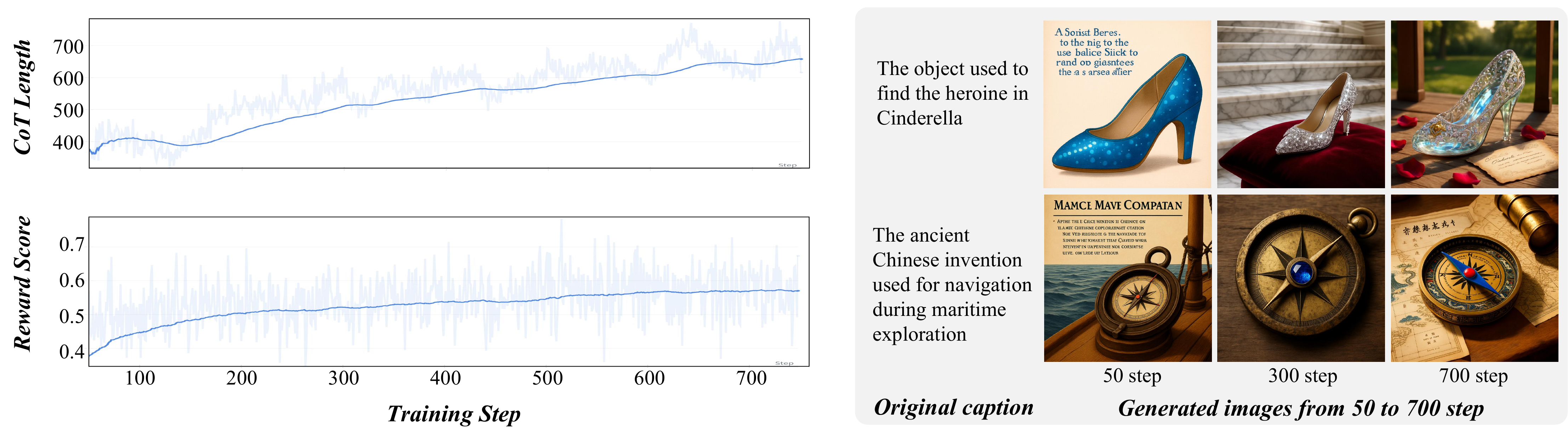}
\end{center}
    \vspace{-6mm}
   \caption{
Visualization of the MLLM-GRPO process, the reward score steadily increases.   
    }
\label{fig:visrl-all}
\end{figure*}

\begin{itemize}[itemsep=2pt,topsep=0pt,parsep=0pt]
\item \textbf{Stage1:} Training only the connector yields inferior text-rendering performance (CVTG: 0.28), indicating insufficient fine-grained alignment between MLLM and DiT.
\item \textbf{Stage2:} Large-scale pre-training results in notable improvements in image quality, with GenEval increasing by 10\%, WISE by 9\%, and CVTG by 35\%.
\item \textbf{Stage3:} High-quality fine-tuning further enhances image details, resulting in an improvement of +12.0\% in CVTG.
\item \textbf{Stage4:} GRPO applied to the MLLM introduces some representation shift in text conditions, slightly affecting image generation on GenEval (-0.01) and WISE (-0.01). However, incorporating CoT significantly boosts reasoning and generation capabilities (WISE: 0.55 $\rightarrow$ 0.76).
\item \textbf{Stage5:} DiT-GRPO further enhances image generation quality, particularly in fine-grained text rendering tasks. (CVTG: 0.79 $\rightarrow$ 0.84)
\end{itemize}


      

\begin{table}[h]
    \scriptsize
    \centering
    \resizebox{0.99\linewidth}{!}{
    \begin{tabular}{l|ccc}
      \Xhline{0.7pt}

      \textbf{Training stage} & \textbf{~~GenEval~~} & \textbf{~~WISE~~} & \textbf{~~CVTG~~}  \\ 
      \hline
      
      Stage1 Alignment  & 0.78 & 0.46 & 0.28 \\     
      Stage2 Pre-training & 0.88 & 0.55 & 0.63 \\    
      Stage3  H.Q. Tuning & 0.88 & 0.55 & 0.75 \\ 
      Stage4 MLLM-GRPO  & 0.86 & 0.54 & 0.75 \\    
      Stage4 MLLM-GRPO & 0.86* & 0.76* & 0.79* \\    
      Stage5 DiT-GRPO & \textbf{0.89}* & \textbf{0.76}* &  \textbf{0.84}*  \\     
      \Xhline{0.7pt}
      \end{tabular}
    }
  \caption{Ablation of training stages of ThinkGen. We use the GenEval, WISE, CVTG for analysis. * denotes that cot reasoning is utilized during image generation.}
      \label{tab:ab-stage}
  \end{table}

\noindent \textbf{Prepadding States.}
We compare the results of Stage1 with and without learnable prepadding states in Tab. \ref{tab:learnable}. Prepadding states significantly improve performance on the \textit{short-prompt benchmarks} 0.64$\rightarrow$0.78 GenEval, 0.37$\rightarrow$0.46 WISEBench, 0.24$\rightarrow$0.28 CVTG and 3.46$\rightarrow$3.93 ImgEdit, indicating that prepadding states can effectively adjust the representation distribution of MLLM output features and promote alignment between MLLM and DiT. 

\begin{table}[h]
    \centering
    \resizebox{0.99\linewidth}{!}{
    \begin{tabular}{l|cccc|c}
    \toprule
     & \multicolumn{4}{c|}{\textit{\textbf{Short-Prompt}}} & \multicolumn{1}{c}{\textit{\textbf{Long-Prompt}}} \\
     & \textbf{GenEval} & \textbf{WISE} & \textbf{CVTG} & \textbf{ImgEdit}  & \textbf{DPG}\\
    \midrule

    w.o. & 0.64 &  0.37 & 0.24 & 3.46 & \textbf{80.90}\\
    w. & \textbf{0.78} &  \textbf{0.46} & \textbf{0.28} & \textbf{3.93} & 80.86 \\
     
    \bottomrule
    \end{tabular}
    }
    \caption{Ablation of the Prepadding States. We divide the evaluation metrics into \textit{long-prompt} (DPG-Bench) and \textit{short-prompt benchmarks} (GenEval, WISE, CVTG, and ImgEdit) for analysis.}
    \label{tab:learnable}
\end{table}

\noindent \textbf{Training strategy.}
In Tab. \ref{tab:training}, we investigate the performance of applying SFT and MLLM-GRPO to the Stage3 model with 10K reasoning data. An interesting phenomenon is observed: directly applying SFT to DiT with reasoning data does not improve performance on reasoning benchmarks, indicating that DiT does not possess the ability to generalize world knowledge to unseen domains. On the other hand, training the MLLM with MLLM-GRPO greatly enhances ThinkGen’s reasoning capability (WISE: 0.55 $\rightarrow$ 0.74).
Therefore, the improvement in ThinkGen’s reasoning generation capabilities is attributable to the SepGRPO training strategy rather than the reasoning data itself.


     

\begin{table}[h]
    \centering
    \resizebox{0.99\linewidth}{!}{
    \begin{tabular}{l|c|lll}
    \toprule
     & \textbf{Training data} & \textbf{GenEval} & \textbf{WISE} & \textbf{CVTG} \\
    \midrule

    \rowcolor{gray! 10} 
    Stage3 & - & 0.88 & 0.55 & 0.75 \\
    SFT & 10K reasoning data & 0.85 \textcolor{red}{$_{-0.03}$} & 0.58 \textcolor{red}{$_{+0.03}$} & 0.67 \textcolor{red}{$_{-0.08}$} \\
    MLLM-GRPO  & 10K reasoning data & 0.80 \textcolor{red}{$_{-0.08}$} & 0.74 \textcolor{red}{$_{+0.19}$} & 0.73 \textcolor{red}{$_{-0.02}$} \\
    MLLM-GRPO  & 24K multitask data & 0.86 \textcolor{red}{$_{-0.02}$} & 0.76 \textcolor{red}{$_{+0.21}$} & 0.79 \textcolor{red}{$_{+0.04}$}  \\
     
    \bottomrule
    \end{tabular}
    }
    \caption{Ablation of the Training strategy. GenEval, WISE and CVTG results are used for analysis. Note: both SFT and Text-GRPO are initialized with the model weights from Stage3.}
    \label{tab:training}
\end{table}

\subsection{Analysis of the SepGRPO Process}
\label{sec:grpo}

We visualize the intermediate process of SepGRPO in Fig. \ref{fig:visrl-all}, including reward scores, CoT length, and generated images. Several key observations emerge: 1) \textbf{Increasing CoT Length:}  The average CoT length gradually grows, suggesting the model develops more sophisticated reasoning during training. 2) \textbf{Unified Reward Growth:} As training progresses, the multi-task reward steadily increases, indicating ThinkGen learns to adaptively think across diverse scenarios. 3) \textbf{Image Quality Improvement:} Visualizations at 50, 300, and 700 steps demonstrate a clear trend of improving image generation quality, with generated images exhibiting richer details and higher fidelity.



\section{Conclusion}
In this work, we introduced ThinkGen, a novel think-driven framework that automatically applies CoT reasoning across diverse generative tasks. Our approach features a decoupled MLLM-DiT architecture trained with SepGRPO, enabling it to formulate a high-quality plan before generation. Extensive experiments demonstrate that ThinkGen achieves significant improvements on reasoning-intensive tasks. Our work represents a key step towards building more intelligent and versatile generative models that seamlessly integrate reasoning and creation.


\clearpage
\maketitlesupplementary

\section{Data Construction}
\label{sec:supp-data}

\subsection{Supervised Training}
\begin{figure}[h]
\begin{center}
   \includegraphics[width=0.99\linewidth]{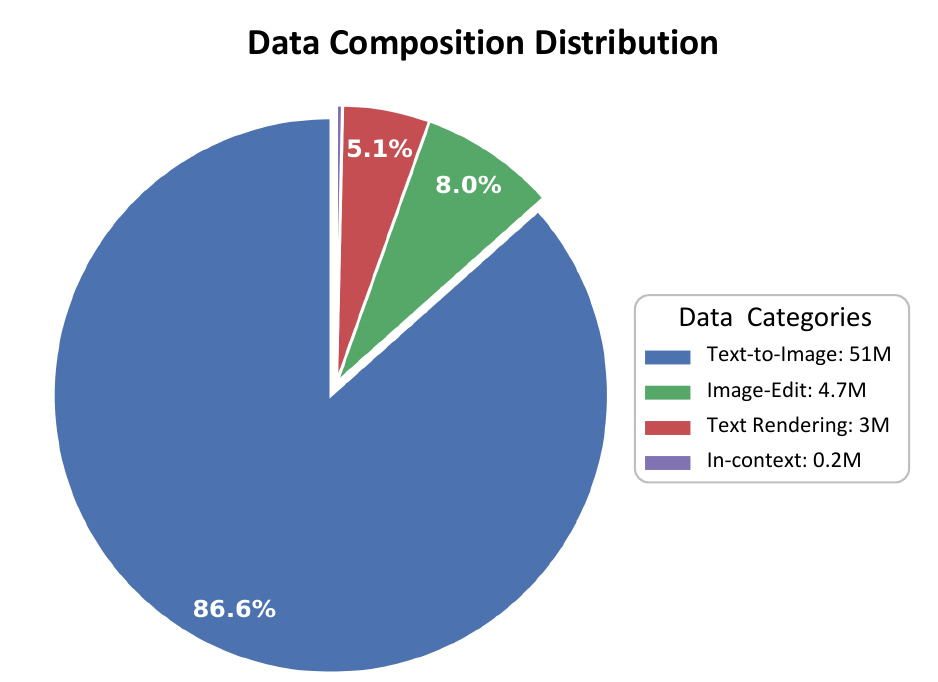}
\end{center}
   \caption{
   Data distribution of supervised training.
    }
\label{fig:data_dis}
\end{figure}

As illustrated in Fig. \ref{fig:data_dis}, we provide an overview of the data distribution utilized for supervised training across different tasks. For the \textbf{text-to-image generation} task, we employ a diverse set of datasets, including ShareGPT-4o-Image \cite{chen2025sharegpt4oimg}, SFHQ \cite{david_beniaguev_2024_SFHQ_T2I}, FLUX-Reason-6M \cite{fang2025flux}, comprising a total of 51M samples. For the \textbf{text rendering} task, we utilize DenseFusion \cite{wang2019densefusion} and internally collected text-containing data, resulting in 3M samples. The \textbf{image editing} task leverages UltraEdit \cite{zhao2024ultraedit}, OmniConsistency \cite{song2025omniconsistency}, Echo4o \cite{ye2025echo4o}, GPT-Image-Edit \cite{wang2025gpt}, ShareGPT-4o-Image \cite{chen2025sharegpt4oimg}, X2Edit \cite{ma2025x2editrevisitingarbitraryinstructionimage}, NHR \cite{Layer2025NoHumansRequired}, accumulating to 5M samples. For \textbf{in-context generation}, we use Nano-banana-150k$^1$\footnote{1. \url{https://github.com/yejy53/Nano-banana-150k}} and Echo-4o-Image \cite{ye2025echo4o}, totaling 200K samples.

\subsection{SepGRPO}

\noindent \textbf{Semantic Composition Dataset.} We employ the Geneval-style training dataset from Flow-GRPO \cite{liu2025flow} as our semantic composition dataset. This dataset comprises prompts that specify object count, color, and relative spatial relationships, making it well-suited for training models to improve semantic alignment between generated images and textual descriptions.

\noindent \textbf{Reasoning Generation Dataset.} 
We collect 10K $\texttt{prompt}$--$\texttt{prompt\_rewrite}$ reasoning data pairs. In each pair, the $\texttt{prompt}$ is intentionally ambiguous and necessitates world knowledge reasoning for text-to-image (T2I) generation, whereas the corresponding $\texttt{prompt\_rewrite}$ is explicit and can be directly used for T2I image generation without further reasoning.
Specifically, we incorporate six types of world knowledge and their respective sub-categories, consistent with the WISE benchmark \cite{niu2025wise}. For each sub-category, we employ GPT \cite{openAI2025gpt4o} to construct $\texttt{prompt}$--$\texttt{prompt\_rewrite}$ pairs (Tab. \ref{tab:supp-temp-wise}).  To ensure the uniqueness, we apply SequenceMatcher for rigorous deduplication, guaranteeing no overlap between our synthesized pairs and the official WISE benchmark.
\begin{table*}[h]\centering
\begin{minipage}{0.9\textwidth}\vspace{0mm}    
    \centering
    \begin{tcolorbox} 
        \centering
        \begin{tabular}{p{0.99\textwidth}}
        \begin{minipage}{0.99\textwidth}
        \texttt{\#\#\#[System Role Instruction]}\\
You are a prompt engineering expert. \\
\\
\texttt{\#\#\#[User Input]}\\
Please generate two prompts for AI image generation. These two prompts must incorporate  $\texttt{sub-category}$ in $\texttt{category}$ knowledge. 

- The first prompt (prompt1) is a more vague prompt that requires $\texttt{sub-category}$ knowledge (this prompt should be as vague as possible and the sentence length should be as short as possible, less than 10 words). For specific writing methods, you can refer to **prompt1** in the Example.\\
- The second prompt (prompt\_rewrite) should provide a straightforward, concrete description of the desired image. This prompt is a clear text-to-image prompt, which can be used to generate images for the text-to-image model without logical reasoning. This T2I\_prompt should be as clear as possible. For specific writing methods, you can refer to **prompt\_rewrite** in the Example)\\
**Return the output as  Do not output anything else.\\
Output only a JSON list, no extra explanation. Strictly generate a list of 5 samples, nothing else. Each sample is a dictionary containing the two keys: ``prompt1'', and ``prompt\_rewrite''.**

        \end{minipage}
        \end{tabular}
    \end{tcolorbox}
    \caption{\textbf{The template to generate reasoning data pairs.}}
    \label{tab:supp-temp-wise}
    \end{minipage}
\end{table*}

\noindent \textbf{Text Rendering Dataset.} We sample 3,000 captions from DataComp-1B \cite{gadre2023datacomp} and employ Qwen3-32B \cite{yang2025qwen3} to rewrite these captions. This rewriting process augments the original descriptions by inserting contextually appropriate text onto specified objects (e.g., placing the word ``coffee'' on a cup).  As a result, the captions are enriched with renderable textual content, making them well-suited for training SepGRPO.

\noindent \textbf{Image Editing Dataset.} We construct our image editing dataset by filtering 3,000 Pico-Banana-400K \cite{qian2025picobanana400klargescaledatasettextguided} samples with near-square aspect ratios (between 0.95 and 1.05). Since both source and target images are resized to square shapes during the MLLM-GRPO training stage, selecting near-square samples helps to minimize distortion caused by resizing. This preprocessing step also facilitates efficient, parallelized reward computation using the SigLIP-2 \cite{tschannen2025siglip}.

\noindent \textbf{Reflection Dataset.} 
We collected 3,000 reflection samples from GenRef-wds \cite{zhuo2025reflection}, a dataset specifically designed for reflection-based image generation. To ensure consistency between images before and after reflection, we exclusively used the \textit{edit} subset in GenRef-wds.

\section{Implementation Details}

ThinkGen integrates Qwen3-VL-8B-Think \cite{qwen3vl} with OmniGen2-DiT-4B \cite{wu2025omnigen2}. The connector is implemented as a simple linear layer that maps the hidden states from the last two layers of Qwen3-VL-8B-Think, reducing their dimensionality from 8,192 to 2,520 to match the input requirements of DiT. For the Prepadding States, we set $K$=25.

As shown in Tab. \ref{tab:supp-imple}, we adopt a multi-stage supervised training strategy using a dynamic mixture of the curated data described in Sec. \ref{sec:supp-data}. Specifically, an alignment stage (Stage1) for initializing the connector, a large-scale pre-training stage (Stage2), and a
supervised fine-tuning stage (Stage3) for high-quality fine-tuning.

During the SepGRPO phase, images are generated at a resolution 512$\times$512 over 20 steps. The $\texttt{cfg}$ parameter is set to 4 and is enabled only during the first 60\% of steps to accelerate generation. The rollout parameters $N_1$ and $N_2$ are set to 8 and 24, respectively. In the DiT-GRPO stage, the loss is backward only for the first 60\% of steps.

\begin{table}[h]
\small
\centering

\begin{tabular}{l|ccc}
\toprule
 & \textbf{Stage1} & \textbf{Stage2} & \textbf{Stage3} \\
\hline
Learning Rate  & $1.0\times10^{-3}$ & $2.5\times10^{-4}$ & $1.0\times10^{-4}$  \\
Batch Size  & 512 & 1280 & 64  \\
LR scheduler   & Cosine & Constant & Constant \\
Weight decay   & 0.0  & 0.0 & 0.0  \\
Gradient Clip  & 1.0 & 1.0 & 1.0 \\
Optimizer       & \multicolumn{3}{c}{AdamW ($\beta_1=0.9$, $\beta_2=0.95$, $\epsilon=10^{-9}$)} \\
Warm-up steps   & 500 & 0 & 0 \\
Training steps  & 47K & 100K & 11k \\
Drop Rate & 10\% & 10\% & 0.01\% \\
Data Size & 24M & 60M & 0.7M \\
Gen resolution  & 512$\times$512 & 512$\times$512 & 1024$\times$1024 \\
\bottomrule
\end{tabular}
\caption{Implementation Details of ThinkGen.} 
\label{tab:supp-imple}
\end{table}

\section{SepGRPO Training Details}

\noindent \textbf{Input Format.} 
During Supervised Pre-training and SepGRPO, we employ distinct data templates for generating pseudo-CoT annotations and for guiding the MLLM in CoT reasoning, as detailed in Sec. 4.1 and Sec. 4.2. Despite their differences, both templates share a common system prompt $\texttt{[SYS]}$ (Tab. \ref{tab:temp-cot-reasoning}), which facilitating a cold start in the RL stages, and encouraging the MLLM to rewrite user input instructions favored by DiT. 

\begin{table*}[!ht]\centering
\begin{minipage}{0.9\textwidth}\vspace{0mm}    
    \centering
    \begin{tcolorbox} 
        \centering
        \begin{tabular}{p{0.99\textwidth}}
        \begin{minipage}{0.99\textwidth}
        \texttt{\#\#\#[System Role Instruction]}\\
You are a helpful, general-purpose AI assistant with the ability to generate images and understand images.

Your primary goal is to assist the user effectively. When generating an image, provide a clear, one-sentence caption that accurately describes the requested image. \\
\\
\texttt{\#\#\#[User Input]}\\
$\texttt{caption}$ or $\texttt{reference images}$ + $\texttt{edit instruction}$

        \end{minipage}
        \end{tabular}
    \end{tcolorbox}
    \caption{\textbf{$\texttt{[SYS]}$ for CoT reasoning.}}
    \label{tab:temp-cot-reasoning}
    \end{minipage}
\end{table*}

\begin{figure*}[t]
\begin{center}
   \includegraphics[width=0.99\linewidth]{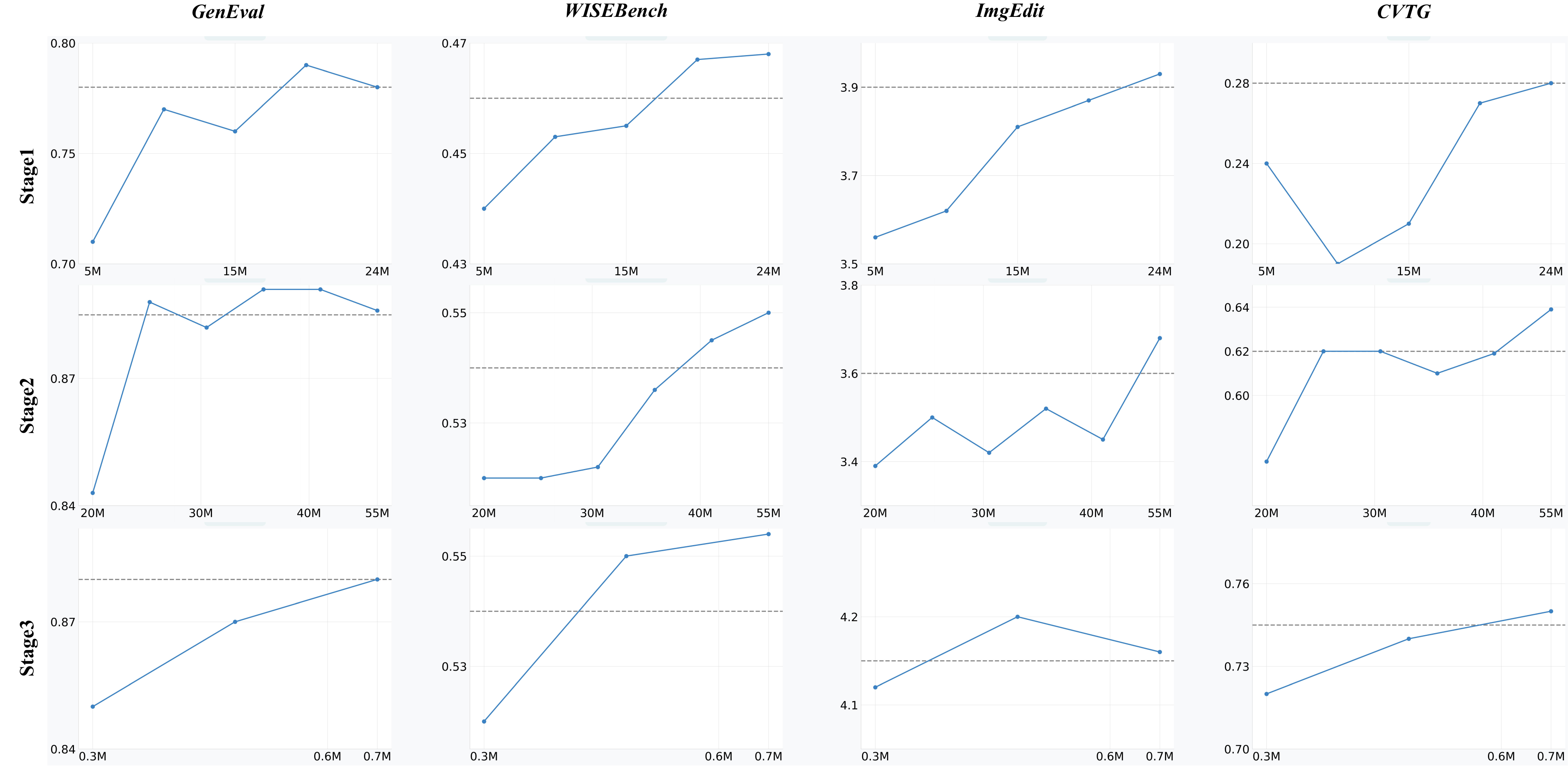}
\end{center}
   \caption{Data Scaling in Stage1-3.
   }
\label{fig:supp_data}
\end{figure*}

\noindent \textbf{Rule Models.} 
SepGRPO employs distinct rule models tailored to each task, as detailed below:
\begin{itemize}[itemsep=2pt,topsep=0pt,parsep=0pt]
\item \textbf{Semantic Composition.} 
We use GenEval \cite{ghosh2023geneval} to evaluate the consistency between generated images and provided instructions.
\item \textbf{Reasoning generation:} 
For this task, images are generated from the $\texttt{prompt}$ in our collected reasoning dataset. The generated image and its corresponding $\texttt{prompt\_rewrite}$ are then scored using HPSv3 \cite{ma2025hpsv3}.
\item \textbf{Text rendering:} 
We utilize 3K prompts containing text rendering. The generated images are processed with OCR \cite{cui2025paddleocr} to extract contained words, and generation quality is assessed via word accuracy \cite{cvtg}.
\item \textbf{Image editing:} 
3K editing samples \cite{qian2025picobanana400klargescaledatasettextguided} are used for CoT reasoning editing. Both the generated images and ground truth are resized to $512\times512$, features are extracted using SigLIP2 \cite{tschannen2025siglip}, and editing quality is measured by cosine similarity.
\item \textbf{Reflection:} 
For this task, 3K reflection samples are split evenly into $\texttt{prompt}$--$\texttt{bad\_image}$ and $\texttt{prompt}$--$\texttt{good\_image}$ pairs. The $\texttt{prompt}$--$\texttt{bad\_image}$ pairs use the corresponding editing instruction as ground-truth, while $\texttt{prompt}$--$\texttt{good\_image}$ pairs use ``\textit{The generated image is well aligned with the caption.}'' as ground-truth. The Normalized Edit Distance (NED) is used to evaluate the MLLM's output. DiT is not used for this evaluation.

\end{itemize}

\section{Supplemental Ablation Study}
In this section, we present ablation studies on connector design and the extraction strategy for the $\texttt{</think>}$ state to validate the effectiveness of our model architecture.

\noindent \textbf{Connector Design.}
Tab. \ref{tab:supp-ab-connector} compares the Stage1 results using different connector designs: a linear layer, a MLP, and a causal-transformer \cite{lin2025uniworld}. The results indicate that the simple linear layer achieves the best performance, outperforming more complex connectors such as MLP and causal-transformer.

\begin{table}[h]
    \scriptsize
    \centering
    \resizebox{0.99\linewidth}{!}{
    \begin{tabular}{l|ccc}
      \Xhline{0.7pt}

      \textbf{Training stage} & \textbf{~~GenEval~~} & \textbf{~~WISE~~} & \textbf{~~ImgEdit~~}  \\ 
      \hline
      
      Linear (default)  & 0.78 & 0.46 & 3.93 \\     
      MLP  & 0.73 & 0.43 & 3.78 \\ 
      Transformer & 0.80 & 0.44 & 3.8 \\    
      \Xhline{0.7pt}
      \end{tabular}
    }
  \caption{Stage1 results of different connector designs. We use GenEval, WISE, ImgEdit for analysis.}
      \label{tab:supp-ab-connector}
  \end{table}

\noindent \textbf{Extraction strategy for the $\texttt{</think>}$ state.}
In VGI-refine block, we truncate the hidden states preceding the $\texttt{</think>}$ token,  feeding only the subsequent hidden states into the DiT. As shown in Tab.~\ref{tab:supp-ab-cut}, this strategy yields consistent improvements across all benchmarks, particularly for \textit{short-prompt} generation tasks (GenEval: +0.12, WISE: +0.15, CVTG: +0.10, ImgEdit: +0.50). These results indicate that truncating the pre-$\texttt{</think>}$ hidden states effectively eliminates redundant information, thereby enhancing image generation quality.

\begin{table}[h]
    \scriptsize
    \centering
    \resizebox{0.99\linewidth}{!}{
    \begin{tabular}{l|cccc|c}
    \toprule
     & \multicolumn{4}{c|}{\textit{\textbf{Short-Prompt}}} & \multicolumn{1}{c}{\textit{\textbf{Long-Prompt}}} \\
     & \textbf{GenEval} & \textbf{WISE} & \textbf{CVTG} & \textbf{ImgEdit}  & \textbf{DPG}\\
    \midrule

    $\texttt{CUT}$ & 0.78 &  0.46 & 0.28 & 3.93 & 80.86 \\
    $\texttt{ALL}$ & 0.66 &  0.31 & 0.18 & 3.43 & 80.60 \\
     
    \bottomrule
    \end{tabular}
    }
  \caption{We analyze the impact of the extraction strategy in VGI-refine using GenEval, WISE, and CVTG. The $\texttt{CUT}$ denotes using only the hidden states following the $\texttt{</think>}$ token for image generation, while $\texttt{ALL}$ employs all hidden states. }
      \label{tab:supp-ab-cut}
  \end{table}

\section{Data Scaling}
We examine the scaling behavior of ThinkGen when increasing training data in Stage1-3 (Fig. \ref{fig:supp_data}).

\begin{figure*}[t]
\begin{center}
    \vspace{-2mm}
   \includegraphics[width=0.99\linewidth]{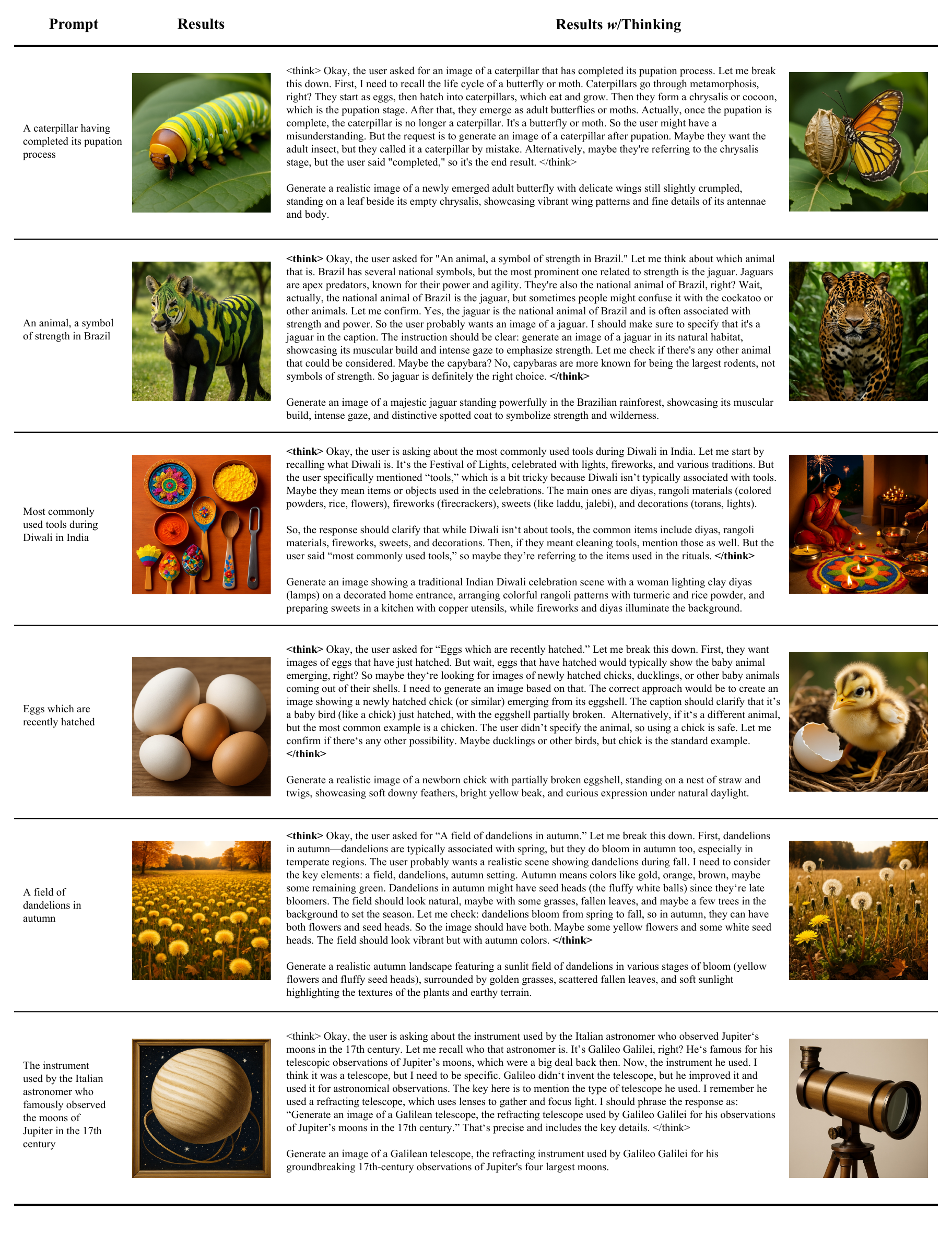}
\end{center}
    \vspace{-6mm}
   \caption{Comparison of image generation with thinking versus without.
   }
\label{fig:supp_vis_gen}
\end{figure*}

\begin{figure*}[t]
\begin{center}
    \vspace{-2mm}
   \includegraphics[width=0.99\linewidth]{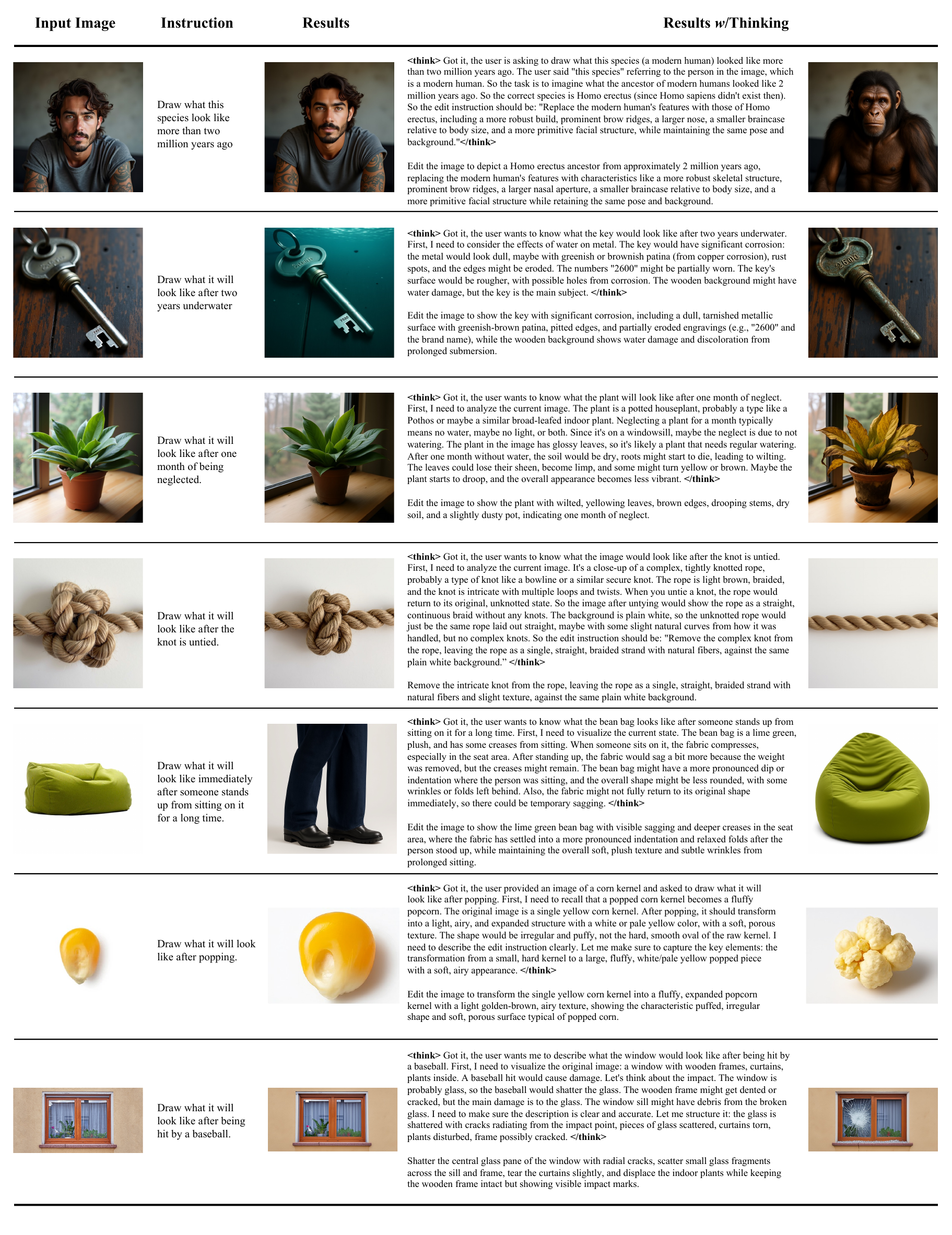}
\end{center}
    \vspace{-6mm}
   \caption{Comparison of image editing with thinking versus without.
   }
\label{fig:supp_vis_edit}
\end{figure*}

\clearpage

{
    \small
    \bibliographystyle{ieeenat_fullname}
    \bibliography{main}
}


\end{document}